# Revolvable Indoor Panoramas Using a Rectified Azimuthal Projection

expanded version of "*An Indoor Alternative to Stereographic Spherical Panoramas*" (Bridges 2014)

Chamberlain Fong
spectralfft@yahoo.com

***Abstract*** – We present an algorithm for converting an indoor spherical panorama into a photograph with a simulated overhead view. The resulting image will have an extremely wide field of view covering up to 4π steradians of the spherical panorama. We argue that our method complements the stereographic projection commonly used in the "little planet" effect. The stereographic projection works well in creating little planets of outdoor scenes; whereas our method is a well-suited counterpart for indoor scenes. The main innovation of our method is the introduction of a novel azimuthal map projection that can smoothly blend between the stereographic projection and the Lambert azimuthal equal-area projection. Our projection has an adjustable parameter that allows one to control and compromise between distortions in shape and distortions in size within the projected panorama. This extra control parameter gives our projection the ability to produce superior results over the stereographic projection.

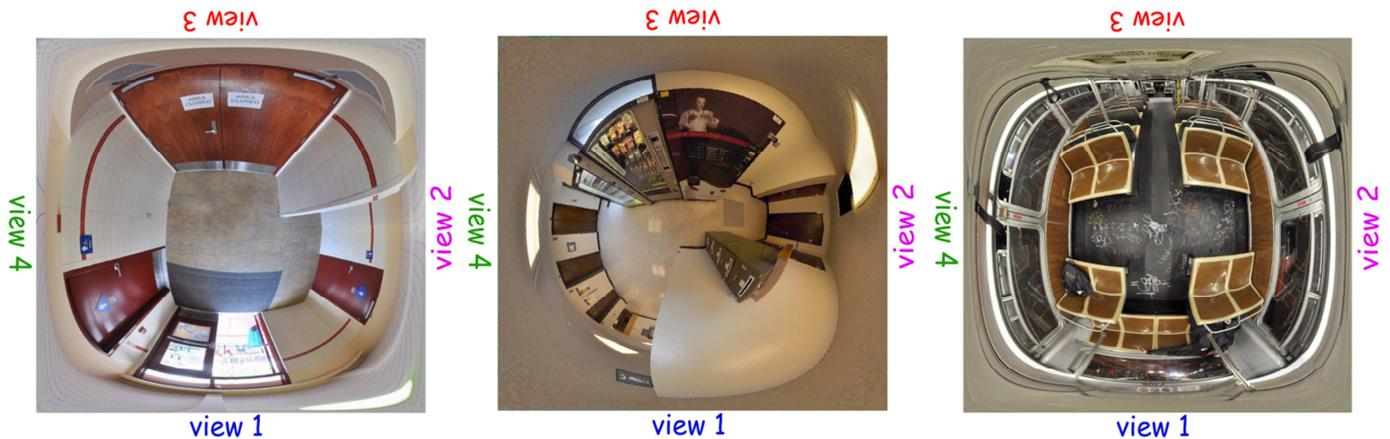

Figure 1: Revolvable panoramas created using our algorithm

## 1 Introduction

Recent advancements in image stitching algorithms and fisheye lens optics have made capturing spherical panoramas easier than ever. Consequently, there are a growing number of photographers who work with such images. Spherical panoramas are the widest possible photographs that one can capture from a single viewpoint. They essentially capture the entire sphere of light that shines over the photographer into a single image.

Many people are familiar with spherical panoramas through the use of Google Street View in conjunction with Google maps. However, one needs an internet-enabled computer in order to view Google's spherical panoramas. Furthermore, Google does not allow users to view its spherical panoramas as a single photograph. It is important to be able to project a spherical panorama into a single photograph which can be viewed statically and printed on a single piece of paper such as a magazine page, a postcard or even a poster.

The stereographic projection is a method for static viewing of spherical panoramas. It is particularly good in producing a fake bird's eye view of an outdoor scene. This effect is commonly known as the "little planet" effect. This moniker came from the stereographic projection's ability to convert spherical panoramas into artistic photographs resembling planetoids in the middle of the sky. This effect is increasingly popular on the internet. In fact, there are several groups in Flickr dedicated to stereographic images. One reason for the popularity of little planets comes from a remarkable property of such images which we shall call *revolvabilty*. Revolvable images exhibit resilience to rotation. That is, if one rotates the image around its center by any angle, one can still get a reasonably intelligible image. In fact, flipping the image upside-down keeps the image just as plausible as the original unrotated version.

In this paper, we will present a spherical map projection derived from blending the stereographic projection with the Lambert azimuthal equal-area projection. This proposed projection can also be considered as a generalization of both projections. Like the stereographic projection, our projection can be used to convert a spherical panorama into a photograph with a simulated overhead view of the scene. In addition, the resulting image will also be revolvable. Unlike the stereographic projection, our projection is suitable for indoor scenes. Figure 1 shows some examples of our results.

## 2 Related Work

Interest in capturing spherical panoramas and applying world map projections to them dates back to Greene's [1986] work in environment mapping. Greene captured spherical panoramas using a camera with 180° FOV fisheye lens and converted them into cube maps to enhance a computer generated scene. Later, Chen [1995] introduced the Apple QuickTime VR system for immersive and interactive viewing of spherical panoramas. Further enhancements to the capture, processing and storage of spherical panoramas were introduced by Debevec [1998], Wan et al. [2007], and Kazhdan et.al. [2010]

German et al. [2007b] discussed the use of different world map projections for static viewing of spherical panoramas. For centuries, mapmakers have studied the problem of representing the spherical Earth on a flat piece of paper [Snyder 1987]. The techniques developed by geographers and cartographers to flatten the earth are also applicable to spherical panoramas.

The equirectangular projection is the most common projection used by the spherical panorama community [Kazhdan et. al 2010]. This is because of its simplicity and ease of use. Equirectangular panoramas are rectangular images with a 2:1 aspect ratio representing a 360°x180° full span of the sphere. In this paper, we will use the equirectangular projection to depict our input spherical panoramas.

There is a burgeoning community of photographers on the internet who specialize in using the stereographic projection for spherical panoramas [German et al. 2007a] [Swart et al. 2011]. The resulting images are often quite compelling because of the wide angle view and revolvability. However, for indoor scenes, the stereographic projection suffers from unnatural and excessive enlargement of features near the ceiling [Fong et al. 2011]. Stereographic projections of indoor panoramas often have a poor balance of size between features in the northern hemisphere and the southern hemisphere. Typically, features near the ceiling and walls are exaggerated and considerably larger than features near the floor.

Stereographic panoramas belong to class of spherical images where one pole of the sphere or some nearby point is at the center of the image. For this paper, we will mostly discuss the case where the South Pole, also known as nadir, is at the center. The nadir of spherical panoramas is usually the ground just beneath the photographer's feet.

The Peirce quincuncial projection is another map projection that has been used to create fake bird's eye views of spherical panoramas. It is a promising projection for producing revolvable overhead views of indoor scenes. However, it has four troublesome non-conformal points that cause unsightly image discontinuities which are difficult to hide [Fong et al. 2011]. Furthermore, the Peirce quincuncial projection has an inherent bias towards the main diagonals of the projected image. It tends to bend straight lines located in regions away from the main diagonals. In addition, the Peirce quincuncial projection is confined to a square. Many rooms are rectangular with a major axis noticeably longer than a minor axis. This often results in unnatural shortening of features in the Peirce quincuncial projection.

## 3 Algorithm Overview

Our algorithm can be summarized in two parts. The 1st part consists of an unoptimized spherical projection that maps the sphere to a square or rectangle. The 2nd part is an optional optimization phase applied to the projection to minimize certain distortion error metrics.

**Phase 1: Unoptimized Spherical Projection**

Our algorithm consists of a map projection that converts the sphere into a square (or rectangle). This map projection has a input parameter $β$ which controls the amount of distortions in the mapping. We shall discuss automated methods for determining the value of this parameter in second part of this overview.

An overview of the pipeline for the map projection is shown in Figure 2. The input to the projection is a spherical panorama. The output is a revolvable panorama with a simulated overhead view of the location.

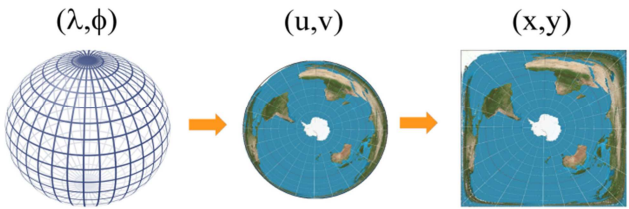

Figure 2: An overview of our projection from input spherical panorama to a revolvable panorama.

The projection consists of 2 steps. The first step is a projection of the sphere onto a circular disc in the plane. We present a novel azimuthal map projection for this step. The second step is to convert this circular disc into a square (or rectangle).

For efficiency reasons, the actual implementation of the projection works backwards by starting from the projected image and fetching pixels from the spherical panorama to fill into the output image. A pseudo-code implementation is shown below. Each step in the pseudo-code corresponds to a box in the block diagram shown in Figure 2, but in reverse order.

---

Unoptimized Spherical Projection algorithm
  Input:   spherical panorama,
           blend parameter $β ∈ (0,1]$
  Output:  revolvable panorama

For each pixel (x,y) in the output image:
  1) Convert the image coordinates (x,y) to corresponding disc coordinates (u,v).
     (See Section 6 for the equations)
  2) Convert the disc coordinates (u,v) to latitude φ and longitude λ on the sphere. This step involves using the inverse equations of our proposed blended azimuthal map projection.
     (See Section 5.2 for the equations)
  3) Fetch the pixel color at the spherical coordinates (λ,φ) of the input panorama and use this as the color value for the pixel (x,y)

---

**Phase 2: Optimization**

This phase of the algorithm consists of finding the optimal value for the blend parameter β to minimize distortions in shape and size within the projection. This step can be skipped and be done manually by an artist if desired.

---

Optimization phase

Find the optimal blend parameter $β ∈ (0,1]$ by
  1) Using a given value of β, compute the spherical projection of the panorama.
  2) Measure the amount of distortion in the projection
     (see Section 7 for the equations)
  3) Adjust the value of β to a different test value depending on search method
  4) Repeat the steps until the measured distortion error can no longer be made smaller

---

## 4 Azimuthal Projections

Azimuthal projections are map projections in which the sphere is projected onto a plane tangent to the sphere at a selected point [Feeman 2002]. This selected point, where the tangent plane intersects with the sphere, will be at the center of the projection. In azimuthal projections, the direction (also known as azimuth) from the center of the projection to every other point on the projection is shown correctly. Moreover, the shortest route from the center to any other point on the projection is a straight line [Snyder 1987]. Thus, azimuthal projections place utmost importance to the center point of the projection. All azimuthal projections map the sphere to a circular disc on a plane, but this disc need not be finite.

Polar azimuthal projections are azimuthal projections that put the North or South Pole at the center of the projection. These projections have many desirable properties that make them particularly useful in the creation of revolvable images. These properties include:

- There is radial symmetry of scale around the center of the projection, which produces naturally circular images;
- Meridians of constant longitude are straight lines emanating radially from the center of the projection;
- Parallels of constant latitude are concentric circles centered at the pole;
- The meridians and the parallels intersect at 90°

When applied to spherical panoramas, polar azimuthal projections produce results that lead to the appeal of revolvable panoramas. Vertical features such as wall corners, posts, and tree trunks are meridians on the sphere. After projection, they remain as straight lines radiating outward from the center of the image. Horizontal features of constant latitude in the spherical panorama are projected to smooth circular arcs. Moreover, these vertical and horizontal features still meet at 90° after projection.

The principal equations for polar azimuthal map projections are:

$$\lambda = \tan^{-1}\left(\frac{u}{v}\right) \qquad \phi = f(r) \text{ where } r = \sqrt{u^2 + v^2}$$

The variables λ and φ are longitude and latitude on the sphere, respectively. The range of values for these geodetic angles is: $-\pi \leq \lambda \leq \pi$ and $-\frac{\pi}{2} \leq \varphi \leq \frac{\pi}{2}$ . The variables u and v are coordinates on the plane after projection to a circular disc. The variable r is the distance of the projection point (u,v) to the center of the disc.

All polar azimuthal projections share closely-related equations for mapping geodetic spherical coordinates (λ,φ) to projective plane coordinates (u,v). In fact, they all have the same expression for longitude λ as $\tan^{-1} u/v$ . Also, the latitude of the projected point only depends on its planar distance $r = \sqrt{u^2 + v^2}$ to the center of the projection. The function *f(r)* can be specified arbitrarily. Each azimuthal projection is distinguished by a different function *f* that expresses latitude in terms of r.

### 4.1 Stereographic Projection

The stereographic projection is an important azimuthal map projection studied and described in Ptolemy's Planisphaerium dating back to 100 A.D. This projection maps the sphere into an infinite plane. The equation for latitude in the south polar aspect of this projection [Snyder 1987][Feeman 2002] is:

$$\phi = 2\,\tan^{-1}(r) - \frac{\pi}{2}$$

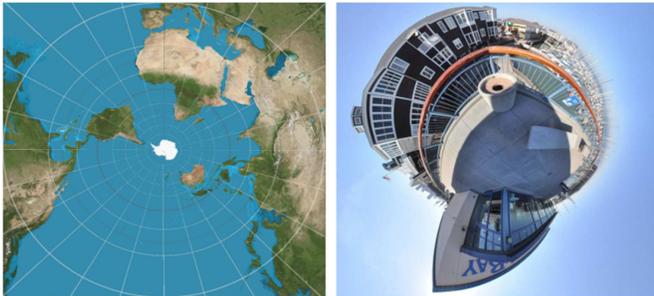

Figure 3: Stereographic projection (left) and an example of panorama with the little planet effect (right). Both images are cropped.

The stereographic projection is a conformal mapping. This means that angles between features are preserved locally after the projection. In other words, small scale shapes are not distorted within the projection. This property makes the stereographic projection useful for photographic applications. In particular, the stereographic projection works especially well in producing "little planets" of outdoor panoramas. It accentuates the shape of features in the upper hemisphere to give a pleasing cartoony effect. However, it deemphasizes the size of features in the lower hemisphere. This is often undesirable for indoor panoramas.

Since the stereographic projection maps the sphere to an infinite plane, cropping is necessary in order to get a finite projection of the spherical panorama. It is possible to get $4\pi - \epsilon$ steradians of spherical coverage using the stereographic projection, where $\epsilon$ is an arbitrarily small solid angle. However, this comes at the expense of extreme enlargement of features near the zenith. The smaller $\epsilon$ gets, the larger the disproportion between the hemispheres will appear in the projected image.

### 4.2 Lambert Azimuthal Equal-Area Projection

Johann Heinrich Lambert developed an important azimuthal map projection in 1772. This projection maps the sphere into a finite circular disc. Figure 4 shows a Lambert azimuthal mapping of the world in its standard form and in a polar aspect.

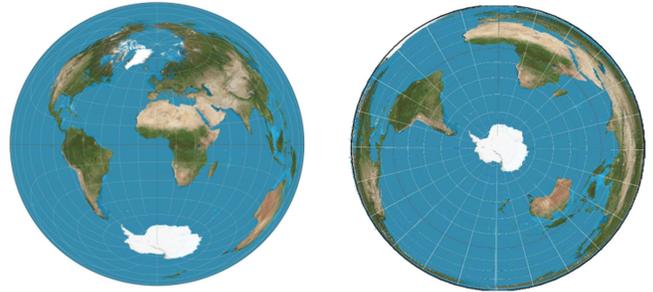

Figure 4: Lambert azimuthal equal-area projection in standard (left) and south polar aspect (right)

In the canonical form of the Lambert azimuthal equal-area projection, the intersection point between the equator and prime meridian is located at the center of the disc. The point at the opposite side of the world is the intersection point between the equator and the international dateline. This antipode point is projected to the whole perimeter of the disc.

The Lambert Azimuthal projection has an important property that makes it useful in many geographic applications. This property is known as the "equal-area" property. In differential geometry parlance, the Lambert azimuthal projection is known as an equiareal projection [Brown 1935]. Equiareal means that the projection preserves the relative size of all features after the mapping. In other words, the area of any feature on the sphere will be proportionally the same to its projected area on Lambert's circular disc. This property is important in keeping a proper balance of size between features in projected panoramas.

Lambert designed his projection so that other spherical aspects are possible. The one with particular interest to us is when the South Pole is at the center of the projection. This is the south polar aspect of the Lambert azimuthal equal-area projection [Snyder 1987] [Cogley 1984]. The equation for latitude in this aspect of the projection is:

$$\phi = 2\,\sin^{-1}(r) - \frac{\pi}{2}$$

This equation holds when the sphere is mapped to an equiareal unit disc on the plane. The distance r from a projected point (u,v) to the center point of the disc is restricted to r≤1. The South Pole (nadir) lies at the center of the circular disc and the North Pole (zenith) is spread across the whole perimeter of the circular disc.

Like the stereographic projection, the Lambert azimuthal projection has its shortcomings when used for projecting to indoor spherical panoramas. Indeed, the Lambert azimuthal projection balances the size of features within the indoor panorama, but this comes at the expense of features appearing unnaturally elongated and squished near the ceiling

## 5 Blended Azimuthal Projection

The stereographic projection is a conformal mapping and the Lambert azimuthal projection is an equiareal mapping. The azimuthal nature of both map projections makes them suitable for creating revolvable panoramas. However, this azimuthal property is usually not enough to make aesthetically-pleasing panoramas. Being conformal or equiareal is also important. Conformal projections preserve angles within the mapping and avert shape distortions in the panorama. Equiareal projections preserve area within the mapping and avert size distortions in the panorama.

Ideally, we want to have a mapping that is both conformal and equiareal. A theorem in differential geometry states that this is equivalent to being an isometry [Kreyszig 1991]. An isometric mapping preserves distances across the entire projection; and in the process, does not distort shape or size. However, for our application of mapping the sphere to the plane, a well-known theorem by Euler in 1775 that states that no such isometric mapping exists [Kreyszig 1991] [Feeman 2002]. In other words, the best that we can do is look for a compromise between being conformal and being equiareal [Van Wijk 2008]. It is impossible to have both properties.

## 5.1 A Blended Compromise

As a compromise, we present an azimuthal projection that essentially blends the stereographic projection with the Lambert azimuthal equal-area projection. We introduce the variable β which acts as blending parameter between the two projections. When β is set to 0, the resulting projection is the stereographic projection. When β is set to 1, the resulting projection is the Lambert azimuthal projection. In between, the projection is a hybrid of the two azimuthal projections. The equation for latitude in this blended azimuthal projection is

$$\phi = 2\tan^{-1}(\frac{r}{\sqrt{1-\beta^2 r^2}}) - \frac{\pi}{2}$$

It is easy to check by substitution and some algebra that the latitude equation for this blended azimuthal projection matches the stereographic equation when β=0. Likewise, it is easy to check that the latitude equation for this blended azimuthal projection matches the Lambert azimuthal equation when β=1 by using the trigonometric identity: $\sin^{-1}(r) = \tan^{-1}(r/\sqrt{1-r^2})$

The stereographic projection maps the sphere to an infinite plane. In contrast, the Lambert azimuthal projection maps the sphere onto a finite circular disc. Needless to say, there is a wide disparity between the span of both projections. In order to have an effective blend of the two projections, we need a projection with a span that can grow from a finite disc to the infinite plane as β goes from 1 to 0. This is exactly what the blended azimuthal projection does – it grows infinitely in size as β approaches 0. In fact, the blended azimuthal projection maps the sphere to a disc with radius $^1/_\beta$

## 5.2 Normalized Form

The vast difference in the spanning range between the stereographic and the Lambert azimuthal projection adds difficulty in creating photographs from blending the two projections. We, therefore, propose a normalized form of the blended azimuthal projection. This normalization can be derived from its unnormalized latitude equation by writing r in terms of a normalized dummy variable defined as $R_{dummy} = r\beta$ , then renaming the dummy variable out of the equation. After which, the equation for latitude becomes:

$$\phi = 2\tan^{-1}(\frac{r}{\beta\sqrt{1-r^2}}) - \frac{\pi}{2}$$

This normalized form of the blended azimuthal projection effectively maps the sphere to a unit disc for all values of β ∈ (0,1]. The only complication with this normalized form is that we are strictly restricted to β>0. That is, this projection cannot be set to 100% stereographic. This limitation stems from the difficulty of linearly mapping an infinitely plane to a unit disc. Nevertheless, β can be set to an arbitrarily small number $\varepsilon > 0$ that can make the projection as close to stereographic as one wishes without actually setting β to zero. This helps us prevent division by zero and other undesirable infinities in the equations. Figure 5 shows the normalized blended azimuthal projection at different values of β from 0.1 to 1. In essence, this is like a sequence of frames of morphing from the nearly stereographic projection to the Lambert azimuthal equal-area projection

In summary, we have presented a blended azimuthal projection that is a hybrid between the stereographic projection and the Lambert azimuthal equal-area projection. Furthermore, we introduced a normalized form that always maps the sphere to a unit disc. These two parameterized continuum of blended projections constitute the *conformal-equiareal spectrum* of azimuthal projections. A table summarizing the key properties of the 4 polar azimuthal projections of interest is provided here.

| Azimuthal projection (south polar aspect) | $\phi = f(r)$ where $r = \sqrt{u^2 + v^2}$ | key property | mapping span | blend value |
|---|---|---|---|---|
| stereographic | $2\tan^{-1}(r) - \frac{\pi}{2}$ | conformal | $0 \leq r < \infty$ | 0 |
| Lambert azimuthal | $2\sin^{-1}(r) - \frac{\pi}{2}$ | equiareal | $0 \leq r \leq 1$ | 1 |
| blended | $2\tan^{-1}(\frac{r}{\sqrt{1-\beta^2 r^2}}) - \frac{\pi}{2}$ | adjustable | $0 \leq r \leq \frac{1}{\beta}$ | β |
| normalized blend | $2\tan^{-1}(\frac{r}{\beta\sqrt{1-r^2}}) - \frac{\pi}{2}$ | adjustable | $0 \leq r \leq 1$ | β |

Table 1: The four polar azimuthal projections used in this paper

The alert reader might notice that the latitude equations given for the stereographic projection and Lambert azimuthal projection are off by a radial factor of 2 from those given in standard map projection text [Snyder 1987]. There is a good reason for this. The surface area of a unit sphere is not the same as the area to a unit disc. Instead, the surface area of a unit sphere equals the area of a circular disc with radius 2. Nevertheless, we want to force our Lambert disc to have unit radius. Likewise, we want the normalized blended azimuthal projection to map to a unit disc; so we are forced to forgo the usual assumption that the input sphere has unit radius. In other words, these equations hold for a sphere of radius ½; or equivalently, unit diameter.

## 5.3 Modifying the Rotational Aspect

In practice, it is often ideal to change the aspect of the spherical projection. That is, one might want to move the center of the projection to a point other than the South Pole. This is useful for placing emphasis or accentuating some features of the spherical panorama. For example, one might want the North Pole to be the center. In this case, it is trivial because one can simply negate the equations for φ. However, usually we want some arbitrary geodetic coordinate (λ₀,φ₀) on the sphere as the center. Moreover, it might even be desirable to change the orientation of the sphere by rotating it about an arbitrary axis by an arbitrary angle. In general, this arbitrary spherical aspect can be represented by a 3D rotation $R_\theta$ on the sphere, where $R_\theta \in \mathbf{SO(3)}$

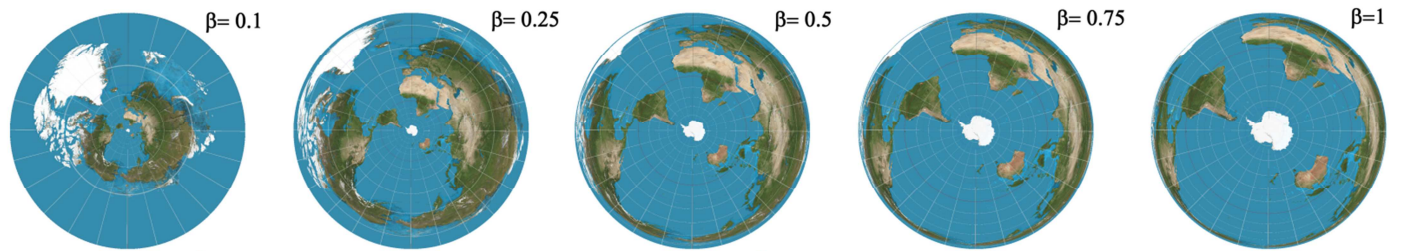

Figure 5: The normalized blended azimuthal projection (south polar aspect) at varying values of *β*

Snyder [1987] kept most of the equations on his map projection treatise as general as possible in order to accommodate different aspects of the projections he covered. However, we believe that this approach make the equations unwieldy and difficult to analyze. Furthermore, some map projections have no known closed-form equations [Cogley 2002]. It is usually more practical to have a canonical aspect of the map projection with key defining equations and then apply the 3D aspect rotation of the sphere as a post-process step [Cogley 1984]. This 3D rotation of the sphere can be done using 3x3 rotation matrices, quaternions, or Euler angles. This technique, effectively, divorces the underlying equations of the projection from the rotational aspect of the sphere. This is the approach that we advocate for calculating different aspects of the blended azimuthal projections discussed in sections 5.1 and 5.2

### 5.4 Inverse Equations

For panoramic imaging applications, we usually just need an expression for spherical coordinate (λ,φ) in terms of planar disc coordinates (u,v). However, there are other applications that require the inverse equations. These inverse equations are:

$$u = r \cos \lambda \qquad v = r \sin \lambda$$

$$for\ blended\ azimuthal: \quad r = \frac{\sin(\frac{\phi}{2} + \frac{\pi}{4})}{\sqrt{1 - (1 - \beta^2) \sin^2(\frac{\phi}{2} + \frac{\pi}{4})}}$$

$$for\ normalized\ case: \quad r = \frac{\beta \sin(\frac{\phi}{2} + \frac{\pi}{4})}{\sqrt{1 - (1 - \beta^2) \sin^2(\frac{\phi}{2} + \frac{\pi}{4})}}$$

Note that the normalized case only holds for **β ∈ (0, 1]**. Recall that the stereographic projection at β=0 is disallowed for the normalized blended azimuthal projection.

## 6 Disc-to-Square Mapping

Most of the world's photographs are rectangular. We are all so accustomed to seeing rectangular photographs that there is a slight psychological aversion to photographs that are not. Besides, rectangles are easily tiled for display in albums, and make much more efficient use of display space than circles and ellipses. This is the main motivation for this extra step.

In this section, we shall introduce a simple algorithm for mapping a circular image to a square. We designed this algorithm to work well with circular azimuthal images as input. In a latter section, we will extend this algorithm to work with ellipses to produce rectangles.

Our problem of mapping a circular disc to a square is similar but not equivalent to the classic mathematical problem of "squaring the circle". For one thing, in the classic mathematical problem, one is restricted to only using a straightedge and a compass. Our problem concerns finding an algorithm that a computer can perform and calculate. So this is a significantly reformulated problem with a specific application of converting circular photographs into square photographs.

The mapping of circular discs to squares has many applications in computer graphics ranging from ray tracing to sampling. Kolb et al. [1995] and Shirley et al. [1997] discussed algorithms applied to such applications. We have found their methods unsuitable for mapping circular images to square images; and hence, came up with a different algorithm.

The canonical space for our mapping is the unit disc centered at the origin inscribed inside a square. This unit disc is defined as $\mathcal{D} = \{(u,v) | u^2 + v^2 \leq 1\}$. Its circumscribing square is defined as $\mathcal{S} = [-1,1] \times [-1,1]$. This square has a side of length 2.

In this paper, we shall denote (u,v) as a point in the interior of the unit disc and (x,y) as the corresponding point in the interior of the square after the mapping. Our goal is to derive an equation that relates (u,v) to (x,y). This equation will ultimately define how the mapping converts a circular disc to a square region. Figure 6 shows a diagram of the unit disc and the square used for the mapping.

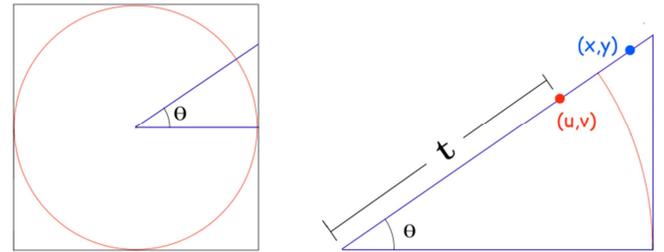

Figure 6: Diagram for the disc-to-square radial mapping process

### 6.1 Radial Constraint

As a design constraint, we impose that the angle that the point (u,v) makes with the x-axis be the same angle as that of point (x,y). We denote this constraint as the radial constraint for the mapping. This effectively forces points to only move radially from the center of the circle during the mapping process. Mathematically, if θ is the angle between the point (u,v) and the x-axis, these equations must hold:

$$\cos \theta = \frac{u}{\sqrt{u^2 + v^2}} = \frac{x}{\sqrt{x^2 + y^2}}$$

$$\sin \theta = \frac{v}{\sqrt{u^2 + v^2}} = \frac{y}{\sqrt{x^2 + y^2}}$$

Meanwhile, each point (u,v) in the interior the circular disc can be parameterized with its polar coordinates as:

$$u = t \cos \theta = t \frac{x}{\sqrt{x^2 + y^2}}$$

$$v = t \sin \theta = t \frac{y}{\sqrt{x^2 + y^2}}$$

where 0 ≤ t ≤ 1 is the point's distance to the origin; and θ is the point's radial angle with the x-axis. The next step is to find a suitable expression for t in terms of x and y; so that we have a mapping equation that relates (u,v) to (x,y).

### 6.2 Fernandez-Guasti's Squircle

Fernandez-Guasti [1992] introduced an algebraic equation for representing an intermediate shape between the circle and the square. His equation included a parameter *s* that can be used to morph from a circle to a square smoothly. This shape has the equation [Weisstein]:

$$x^2 + y^2 - \frac{s^2}{k^2} x^2 y^2 = k^2$$

The parameter *s* can have any value between 0 and 1. When s=0, the equation produces a circle with radius k. When s=1, the equation produces a square with a side length of 2k. In between, the equation produces a smooth curve that resembles both shapes.

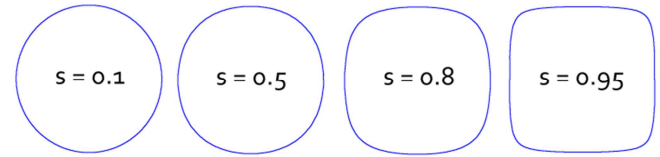

Figure 7: FG-squircle with varying s parameter values

Using the squircle, we can design a way to map a circular photograph smoothly to a square photograph. The main idea is to map each circular contour in the interior of the disc to a squircle in the interior of the square.

In order to get a continuum of growing concentric squircles in the interior of the square, we impose a simple rule s = k in Fernandez-Guasti's squircle equation. This effectively reduces the squircle equation to $x^2 + y^2 - x^2y^2 = s^2$. Furthermore, by varying s from 0 to 1, we get contour curves that fill the interior of a square with concentric squircles growing in size. We then assign each squircular contour of parameter s inside the square to a circular contour of parameter t inside the circular disc. This is done by setting $s = t = \sqrt{x^2 + y^2 - x^2y^2}$. Substituting back, we get a simple equation relating the point (u,v) on the disc to the point (x,y) on the square.

$$u = \frac{x\sqrt{x^2 + y^2 - x^2y^2}}{\sqrt{x^2 + y^2}} \qquad v = \frac{y\sqrt{x^2 + y^2 - x^2y^2}}{\sqrt{x^2 + y^2}}$$

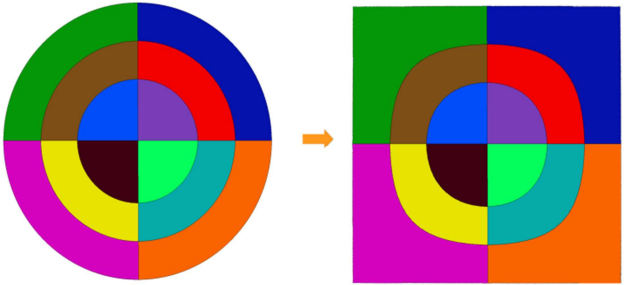

Figure 8: A circular disc input pattern (left) and its corresponding square pattern (right) after using the FG-squircular mapping.

We shall denote this mapping between disc and the square as the FG-squircular mapping. This mapping is radially constrained. There is a good reason for this. Using the radial constraint in our mapping allows us to carry over many of the desirable properties of the polar azimuthal projection to the square image. For example, the meridians remain as straights lines emanating radially from the center of the image after the mapping.

### 6.3 Other Disc-to-Square Mappings

The FG-squircular mapping that we presented here is not the only way to map a disc to a square. In fact, there are many different ways to map the disc to a square and vice versa [Fong 2014]. Moreover, the FG-squircular mapping is neither conformal nor equiareal. This means that it distorts both shape and area when converting the disc to a square. There exist other disc-to-square mappings that are equiareal; and others that are conformal. For example, the concentric map by Shirley et al. [1997] is equiareal.

A well-known method for mapping the circular disc to a square region is through the use of the Schwarz-Christoffel transformation from complex analysis. In its general form, the Schwarz-Christoffel transformation maps the circle to an n-sided polygon conformally. A special case of this maps the circle to a square conformally. The Peirce quincuncial projection [Fong et. al 2011] is an example of a map projection that uses the Schwarz-Christoffel conformal mapping as part of the process. For a more thorough discussion of mappings between the square and the circular disc including forward and inverse equations as well as proofs and derivations, see [Fong 2014]

## 7 Adjusting the Blend Parameter

The simplest way to decide on a value for the blend parameter β of our rectified azimuthal projection is to do it manually. Figure 9 shows the effect of varying the blend parameter β for the rectified azimuthal projection of a panorama for indoor and outdoor scenes. Typically, an artist can adjust β manually and evaluate the aesthetic quality of the resulting projection. However, this can be laborious so we will discuss automated ways of calculating the value of β in this section.

Ideally, we want to find a blend parameter so that the projection produces both a conformal and equiareal mapping of the sphere onto the plane. However, as previously mentioned [Kreyszig 1991], this is impossible because the sphere is a non-developable surface. We will have to settle for a value of β that minimizes conformal error and equiareal error on the mapping. Moreover, there has to be some sort of compromise between the two.

Floater and Hormann [2005] used techniques from differential geometry to write an elegant and succinct methodology for measuring conformal and equiareal errors in mappings. We will briefly summarize their methods and write down equations for calculating error metrics in our projection.

The first step is to write down the equations for relating the point $(X_3,Y_3,Z_3)$ on the sphere to its corresponding point (x,y) on the projected plane. Specifically, we look for 3 functions $f_1$, $f_2$, $f_3$ of x & y such that: $X_3 = f_1(x,y) \qquad Y_3 = f_2(x,y) \qquad Z_3 = f_3(x,y)$

This is essentially the merging of several equations in the projection pipeline. Recall that we have equations for converting a 3-D point on the sphere $(X_3,Y_3,Z_3)$ to geodetic spherical coordinates (λ,φ) then to circular disc coordinates (u,v), then to square coordinates (x,y) on the plane. Using algebra and substituting variables in this chain of equations, we can get simplified expressions for $f_1$, $f_2$, $f_3$:

$$f_1(x,y) = \frac{\beta y\sqrt{x^2 + y^2 - x^2y^2}\sqrt{(x^2-1)(y^2-1)}}{\sqrt{x^2+y^2}[\beta^2 + (1-\beta^2)(x^2+y^2-x^2y^2)]}$$

$$f_2(x,y) = \frac{\beta x\sqrt{x^2 + y^2 - x^2y^2}\sqrt{(x^2-1)(y^2-1)}}{\sqrt{x^2+y^2}[\beta^2 + (1-\beta^2)(x^2+y^2-x^2y^2)]}$$

$$f_3(x,y) = \frac{(1+\beta^2)(x^2+y^2-x^2y^2) - \beta^2}{2(1-\beta^2)(x^2+y^2-x^2y^2) + 2\beta^2}$$

The next step is to use a concept in differential geometry known as the first fundamental form [Kuhnel 2006]. This is a 2x2 matrix:

$$I_f = \begin{pmatrix} E & F \\ F & G \end{pmatrix} \quad where$$

$$E = \left(\frac{\partial f_1}{\partial x}\right)^2 + \left(\frac{\partial f_2}{\partial x}\right)^2 + \left(\frac{\partial f_3}{\partial x}\right)^2 \qquad G = \left(\frac{\partial f_1}{\partial y}\right)^2 + \left(\frac{\partial f_2}{\partial y}\right)^2 + \left(\frac{\partial f_3}{\partial y}\right)^2$$

$$F = \left(\frac{\partial f_1}{\partial x}\right)\left(\frac{\partial f_1}{\partial y}\right) + \left(\frac{\partial f_2}{\partial x}\right)\left(\frac{\partial f_2}{\partial y}\right) + \left(\frac{\partial f_3}{\partial x}\right)\left(\frac{\partial f_3}{\partial y}\right)$$

The first fundamental form is a Jacobian matrix of the transformation from $(X_3,Y_3,Z_3)$ on the sphere to (x,y) on the square. The derivatives needed for the E, F, and G terms can be calculated using methods from numerical differentiation or by using calculus to work out long symbolic equations that can be numerically evaluated.

The first fundamental form is a symmetric and positive definite matrix for which one can perform singular value decomposition to get singular values $\sigma_1$ and $\sigma_2$. These singular values can then be used to measure the amount of distortion on the mapping. The characteristic values [Floater et.al 2005] of $\sigma_1$ and $\sigma_2$ are summarized in the following table:

| Mapping type | preserves | $I_f$ singular values | $\sigma_1$ |
|---|---|---|---|
| Isometry | length | $\sigma_1 = \sigma_2 = 1$ | 1 |
| Conformal | angle | $\sigma_1 = \sigma_2$ | $\sigma_2$ |
| Equiareal | area | $\sigma_1 \sigma_2 = 1$ | $1/\sigma_2$ |

Note that the convention for singular values is $\sigma_1 \geq \sigma_2 \geq 0$. The values of $\sigma_1$ and $\sigma_2$ vary for every pixel in the panorama. They also depend on the value of the blending parameter β. Ideally, we want $\sigma_1 = \sigma_2 = 1$ for every pixel in the projection. This criterion would make the projection an isometry, which is both conformal and equiareal. However, this is impossible for spherical mappings on the plane. Instead, we shall use metrics that quantify how much $\sigma_1$ and $\sigma_2$ deviate from prescribed values given in the characteristic table shown above. These metrics allow us to measure the conformal

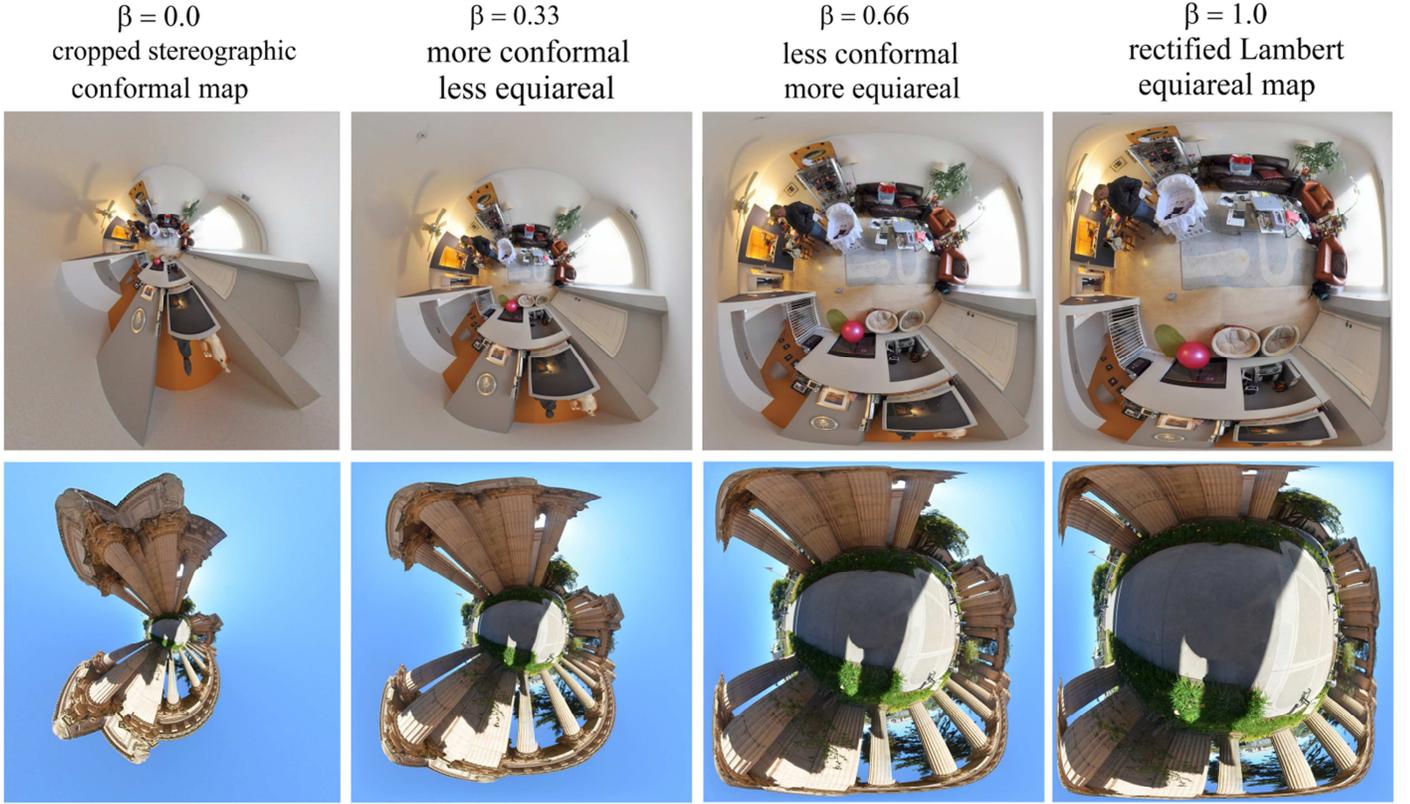
Figure 9: Adjusting the blend parameter of the rectified azimuthal projection

error and equiareal error of our projection for a given blend parameter β. First, we define the product and ratio of our singular values as $\sigma_{product}$ and $\sigma_{ratio}$ respectively. However, in order to avoid undesirable infinities in our equations when $\sigma_1$ or $\sigma_2$ are zero, we limit $\sigma_{product}$ and $\sigma_{ratio}$ to the interval [0,1] by using these equations:

$$\sigma_{product} = argmin\left(\sigma_1\sigma_2, \frac{1}{\sigma_1\sigma_2}\right) \qquad \sigma_{ratio} = argmin(\frac{\sigma_2}{\sigma_1}, 1)$$

We can then define error metrics $e_c$ and $e_q$, which are the conformal error and equiareal error, respectively, as:

$$e_c = (1 - \sigma_{ratio}) \qquad e_q = (1 - \sigma_{product})$$

Both the conformal error metric $e_c$, and the equiareal error metric $e_q$ vary from pixel to pixel in the projected panorama. If the projection is purely conformal (e.g. the stereographic projection), $e_c$ will be zero for every pixel of the panorama. Likewise, if the projection is purely equiareal (e.g. the Lambert azimuthal projection), $e_q$ will be zero for every pixel of the panorama. In general, for the blended rectified azimuthal projection introduced in this paper, $e_c$ and $e_q$ will have varying non-zero positive values for most pixels, depending on the blend parameter β. In order to measure the total distortion error of the projection for a given blend parameter β, one needs to compute a weighted sum of $e_c$ and $e_q$ for every pixel in the panorama.

When measuring distortion, it is important to only measure conformal and equiareal errors on pixels that matter. That is to say, we do not care about shape and size distortions in pixels with low-energy. For example, the sky in stereographic panoramas consists of mostly homogenous pixels with not much features. These pixels can be ignored or even cropped without significantly affecting the image content. In general, we want to bias the conformal and equiareal error measurements towards pixels with high-saliency.

There are many proposed image saliency metrics in the graphics literature. For our algorithm, we have found that using the $e_1$ energy function [Avidan et. al 2007] as adequate. This energy function for measuring pixel saliency in the input image $\mathbf{I_m}$ is defined as the $L_1$-norm of the image gradient.

$$e_1 = \left|\frac{\partial}{\partial x}\mathbf{I_m}\right| + \left|\frac{\partial}{\partial y}\mathbf{I_m}\right|$$

Using this image saliency metric, we can define the total distortion error on our blended rectified azimuthal projection as

$$e_{total} = \sum_{all\ pixels} e_1(k_c e_c + k_q e_q)$$

where $k_c$ and $k_q$ are constant weight factors that introduce partiality of importance between conformal errors and equiareal errors. If one places equal importance to being conformal and being equiareal, $k_c$ and $k_q$ can be set to 1. In practice, we have found that conformal distortion artifacts are slightly more undesirable than equiareal distortion artifacts, so we put more weight on conformal errors.

Given this method for measuring total distortion error of our blended rectified azimuthal projection, it is now possible to search for an optimal blend parameter β that minimizes this error. The search range for β is within the small interval (0,1]. The total distortion error metric is image dependent because it emphasizes the measurement of distortion errors in pixels with high saliency. This means that the optimal blend parameter β varies for different images.

## 8 Elliptical Extension

We now generalize our method in order to be able to produce rectangular photographs instead of square photographs. This modification allows us to produce more convincing overhead views of rectangular rooms, since many indoor rooms do not have walls of equal length on all 4 sides.

**8.1 Elliptical Form of Blended Azimuthal Projection**

We first generalize the blended azimuthal projection to map the sphere to an ellipse instead of a circle. The equations for calculating latitude and longitude are:

$$\lambda = \tan^{-1}(\frac{au}{bv}) \qquad \phi = \tan^{-1}(\frac{r}{\beta\sqrt{1-r^2}}) - \frac{\pi}{2}$$

$$\text{for the ellipse: } r = \sqrt{\frac{u^2}{a^2} + \frac{v^2}{b^2}}$$

where *a* and *b* are the lengths of the semi-major and semi-minor axes of the ellipse, respectively. For a normalized ellipse, we set b=1 to force the vertical axis to be of unit length. The horizontal length can vary depending on the variable *a* to produce different eccentricities of the ellipse

### 8.2 Ellipse-to-Rectangle Mapping

We also extend the disc-to-square mapping algorithm described in a previous section to map ellipses to rectangles. The main idea of our extension here is to scale down the input ellipse into a circle, map this circle into a square, then undo the scaling to get a rectangle. Let the input ellipse have a semi-major length *a* and a semi-minor length *b*. Given functions **g** and **h** previously defined in Section 6 for disc-to-square mapping; i.e. *u=**g**(x,y)* and *v=**h**(x,y)*, the corresponding equations for the ellipse are

$$u = a\, \mathbf{g}\left(\frac{x}{a}, \frac{y}{b}\right) \qquad v = b\, \mathbf{h}\left(\frac{x}{a}, \frac{y}{b}\right)$$

Specifically, for the FG-squircular mapping defined in Section 6.2, these equations become:

$$u = \frac{ax\sqrt{\frac{x^2}{a^2} + \frac{y^2}{b^2} - \frac{x^2 y^2}{a^2 b^2}}}{\sqrt{\frac{x^2}{a^2} + \frac{y^2}{b^2}}} \qquad v = \frac{by\sqrt{\frac{x^2}{a^2} + \frac{y^2}{b^2} - \frac{x^2 y^2}{a^2 b^2}}}{\sqrt{\frac{x^2}{a^2} + \frac{y^2}{b^2}}}$$

## 9 Results

We show images produced by our projection applied to several equirectangular images in the top section of the results page. Most of the images follow a basic layout which we call the fundamental floor plan. This floor plan is illustrated in Figure 10.

We also show elliptically-extended panoramas useful for visualizing rectangular rooms in the third row. Such examples show that our projection need not be limited to a square.

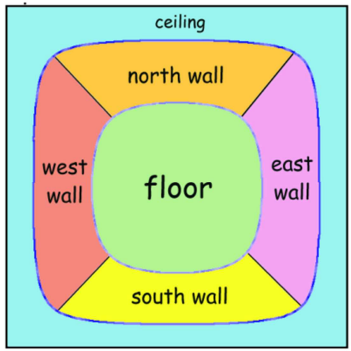

Figure 10: The fundamental floor plan given by rectified azimuthal projections of indoor scenes.

### 9.1 Uncapping the Ceiling

The ceiling is usually a problem in revolvable indoor panoramas. In the case of the stereographic projection, the ceiling is unnaturally enlarged because the stereographic projection is not equiareal. Furthermore, cropping is mandatory in order to get a finite image. In contrast, the Lambert azimuthal equal-area projection has unnatural squishing and elongation of features near the ceiling. This is because the Lambert azimuthal projection is not conformal.

The ceiling's antipode on the sphere is the floor, which maps to the center of the revolvable panorama. Using the distortion error metrics given in Section 7, one can see that both the cropped stereographic and the Lambert azimuthal equal-area projections are actually close to being simultaneously conformal and equiareal in this region.

Blending the stereographic and the Lambert azimuthal projections does not completely get rid of all ceiling distortion problems. It only gives one the ability to compromise between size and shape distortions. Hence, sometimes the best way to generate an aesthetically-pleasing revolvable indoor panorama is to simply crop out the ceiling. This essentially produces a panorama consisting of the room's 4 walls with the floor at the center. This approach still offers an excellent way for visualizing indoor rooms. It essentially provides the photographer the ability to shoot at the ground level and later post-process the pictures to get a fake bird's eye view of the scene.

### 9.2 Outdoor Panoramas

Although we emphasized the use of our algorithm for indoor scenes in this paper, we would like to mention that our method also works with outdoor scenes. Figure 9 shows a progression of the rectified azimuthal projection for an outdoor scene with β varying from 0 to 1. The cropped stereographic little planet (leftmost image) highlights one of the inherent problems with the stereographic projection. Whenever there are very tall features in the outdoor panorama, such as the monolithic columns in our example, there is excessive enlargement of features near the zenith of the panorama. Moreover, this enlargement comes at the expense of the other features within the panorama, specifically those near the nadir which get reduced in size. In our example, the nadir region is shrunk to the point of being barely perceptible. In contrast, the Lambert azimuthal equal-area panorama at the right tends to squish features near the zenith, which effectively makes shapes difficult to discern. The two middle images offer a compromise between the stereographic and the Lambert azimuthal projections; and give results that balance distortions in size and shape.

## 10 Discussion

We believe that the key ingredient to the appeal of the revolvable panoramas such as those used in the "little planet" effect comes from the azimuthal nature of the projection. We, therefore, discuss other alternative azimuthal projections here. There are many other azimuthal map projections with polar aspects that can be used for creating revolvable panoramas. The most notable of these are the orthographic projection and the gnomonic projection. However, both of these projections suffer a severe drawback of not being wide enough to represent the full spherical panorama. Both of these projections can only cover at most $2\pi$ steradians of the sphere; or just a hemisphere. On the other hand, both projections can give realistic overhead views of the spherical panorama. In fact, since the gnomonic projection is the mapping used by some rectilinear photographic lenses, the gnomonic projection can provide results similar to real overhead cameras with ultra wide angle lens. However, this is not necessarily ideal for our application of visualizing spherical panoramas in full.

Another azimuthal projection of interest is the *azimuthal equidistant* projection. This projection actually performs relatively well in terms of angle distortion and area distortion metrics [Feeman 2002] [Floater 2005] and can produce aesthetically-pleasing revolvable panoramas. Debevec [1998] used this projection in representing his HDR panoramas for image-based lighting. However, unlike the blended azimuthal projection, the azimuthal equidistant projection does not give the user control over the compromise between being conformal and being equiareal. Hence, it produces revolvable panoramas that are less flexible and inferior to our blended azimuthal method.

Our blended azimuthal projection is by no means the first map projection to attempt to hybridize the stereographic and the Lambert azimuthal equal-area projection. Snyder [1997] mentions at least three projections that do so: the Breusing Geometric Mean azimuthal projection, the Breusing Harmonic Mean azimuthal projection, and the closely related Airy Minimum-Error azimuthal projection [Yang 1999]. However, as mentioned in the previous paragraph, our projection is more flexible than these projections; and gives the user control over the compromise between size and shape distortions.

## 11 Summary and Conclusion

We presented the use of a rectified azimuthal projection for creating revolvable panoramas with a simulated overhead view of indoor scenes. Our method is an alternative to the stereographic and Peirce quincuncial projections for visualizing spherical panoramas. The main innovation of our proposed technique is the use of a blended azimuthal projection in conjunction with a novel disc-to-square mapping. . Finally, the main message of this paper is to convey the importance of the compromise between being equiareal and being conformal in spherical panorama projections. Conformality is simply not good enough for indoor scenes!

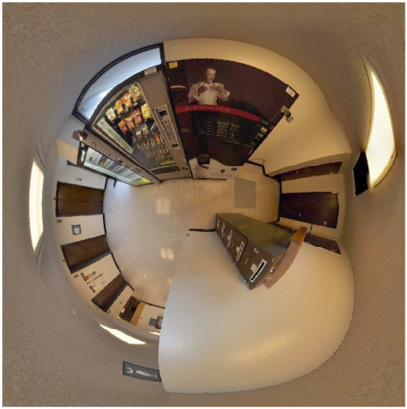
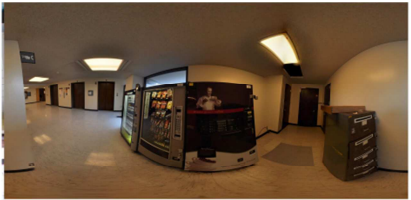
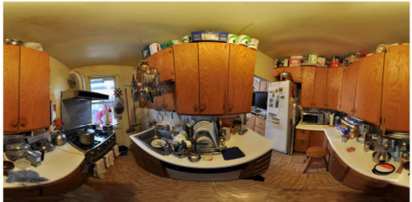
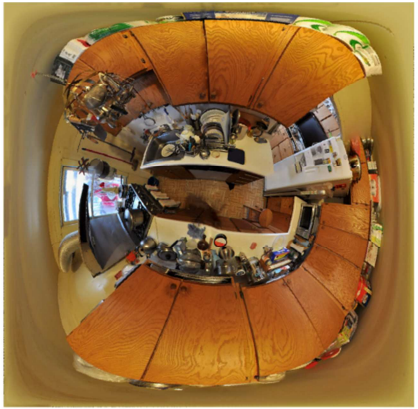
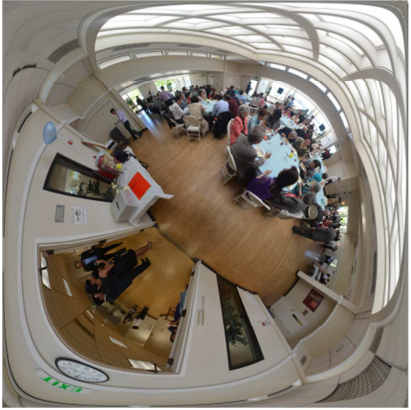
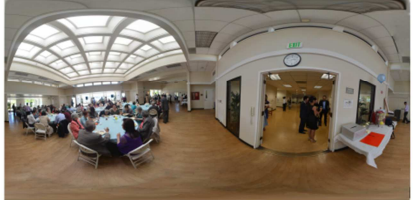
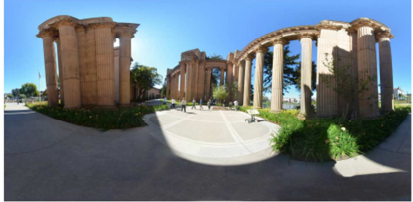
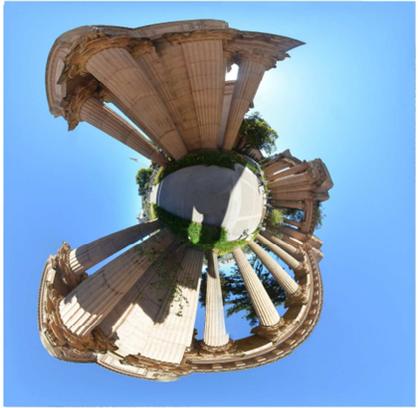
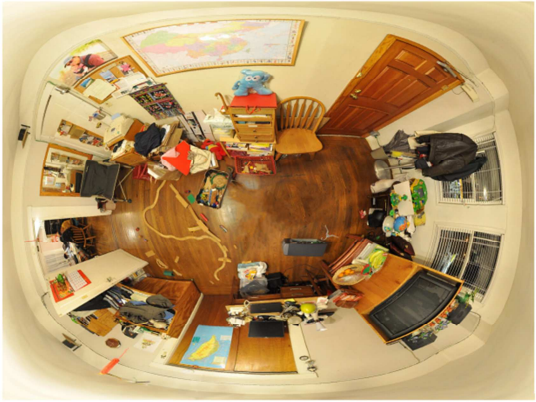
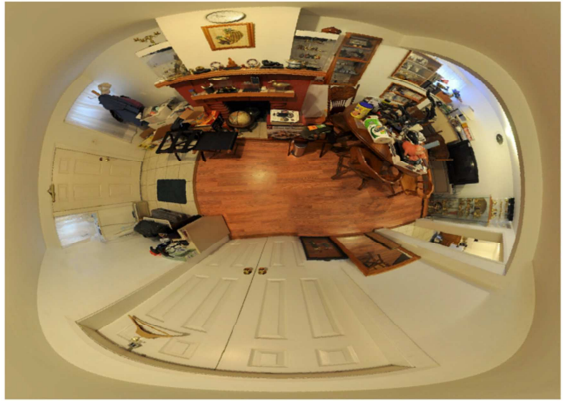
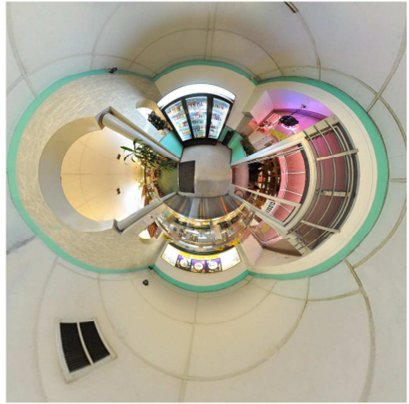
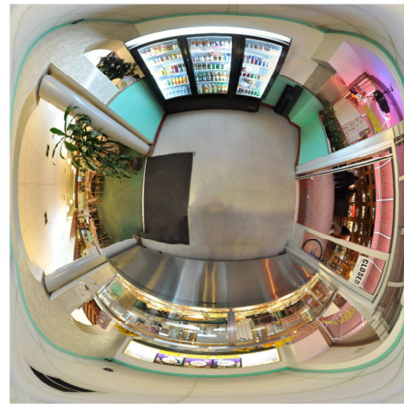
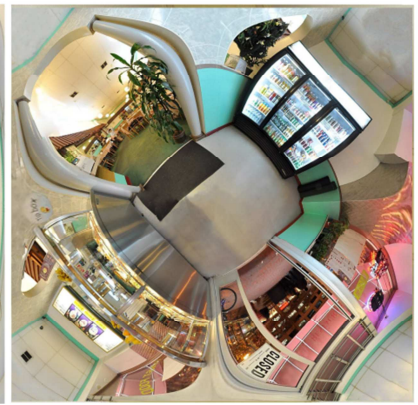

# Appendix 1: Other Disc-to-Square Mappings

In Section 6 of this paper, we presented a simple way to map every point inside the circular disc to a point inside the square. The mapping we presented is in no way unique. There are many different ways to map a circular disc to a square region. In this appendix, we will discuss some other disc-to-square mappings.

Recall the definitions given in section 6. We map a point (x,y) inside the unit disc $\mathcal{D} = \{(u,v) | u^2 + v^2 \leq 1\}$. to a point (u,v) inside the disc's circumscribing square $\mathcal{S} = [-1,1] \times [-1,1]$. . In order to define a mapping, we need to provide equation that relates (u,v) to (x,y). In this appendix, we will present some other disc-to-square mappings. We shall limit these mappings to obey the radial constraint discussed in Section 6.1. That is, these mappings follow these equations for some expression t dependent on x and y:

$$u = t \frac{x}{\sqrt{x^2 + y^2}}$$

$$v = t \frac{y}{\sqrt{x^2 + y^2}}$$

## A1.1 Simple Stretching

One of the simplest ways to perform a disc-to-square mapping is to linearly stretch the circle to the rim of the circumscribing square. The equations for stretching from rim to rim are simple but needs to be broken down to four different cases.

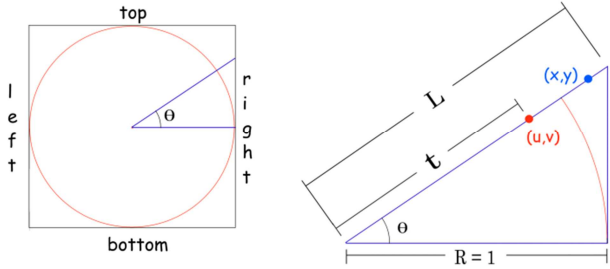

Figure A1.1: Simple Stretching diagram (left) with 4 labeled walls and a zoomed triangle (right) depicting the case for the wall on the right

We first consider the case where the circle extends to the east wall. This occurs for angle θ such that $-45° \leq \theta \leq 45°$ . If we parameterize t to be linearly proportional to the distance of the destination point (x,y) from the origin, we get:

$$\frac{t}{R} = \frac{\sqrt{x^2 + y^2}}{L}$$

Note that R=1 for our unit circle. Using trigonometry we have $\cos \theta = 1/L$ ,hence $t = \sqrt{x^2 + y^2} \cos \theta$. Also from trigonometry, we have $\cos \theta = \frac{x}{\sqrt{x^2+y^2}}$ so, the equation simplifies to $t = x$ for the east wall. Using the same reasoning, we can get the value of $t$ for the other walls

$$t = \begin{cases} x, & \text{for the right wall} \quad \leftrightarrow \quad x \geq |y| \\ y, & \text{for the top wall} \quad \leftrightarrow \quad |x| \leq y \\ -x, & \text{for the left wall} \quad \leftrightarrow \quad -x \geq |y| \\ -y, & \text{for the bottom wall} \quad \leftrightarrow \quad |x| \leq -y \end{cases}$$

Substituting back into the radial constraint equation described in Section 6.1, we get an equation that relates the point (u,v) in the circular disc to its corresponding point (x,y) in the square.

$$u = \begin{cases} \frac{x^2}{\sqrt{x^2 + y^2}}, & \text{for the right wall} \\ \frac{x y}{\sqrt{x^2 + y^2}}, & \text{for the top wall} \\ \frac{-x^2}{\sqrt{x^2 + y^2}}, & \text{for the left wall} \\ \frac{-x y}{\sqrt{x^2 + y^2}}, & \text{for the bottom wall} \end{cases}$$

$$v = \begin{cases} \frac{x y}{\sqrt{x^2 + y^2}}, & \text{for the right wall} \\ \frac{y^2}{\sqrt{x^2 + y^2}}, & \text{for the top wall} \\ \frac{-x y}{\sqrt{x^2 + y^2}}, & \text{for the left wall} \\ \frac{-y^2}{\sqrt{x^2 + y^2}}, & \text{for the bottom wall} \end{cases}$$

At first glance, the Simple Stretching map gives results resembling the Shirley-Chiu concentric map [Shirley et al. 1997]. However, this similarity is only superficial. For one thing, the Shirley-Chiu concentric map does not satisfy the radial constraint equations given in Section 6.1. Also, the Simple Stretching map is not equiareal, but the Shirley-Chiu concentric map is equiareal.

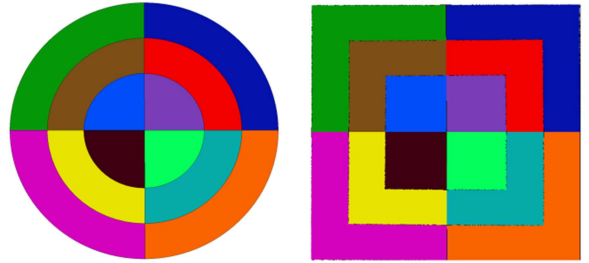

Figure A1.2: A circular disc input pattern (left) and its corresponding Simple Stretching map to a square. There are bend discontinuities along the main diagonals of the square. These bend discontinuities are usually undesirable for photographs.

The Simple Stretching map does share a key qualitative characteristic as the Shirley-Chiu concentric map. Both map concentric circles from the unit disc to concentric squares in the inscribing square. For imaging applications, this is not quite ideal because it causes unwanted bend discontinuities in the resulting image.

## A1.2 Blending the Simple Stretching map

By itself, the Simple Stretching map gives poor results. However, using an idea proposed by Bedard [2009], it is possible to blend this mapping with the original input image to get good results. The main idea is to use linear interpolation to blend between the input disc and its mapped square region to produce a method with less-pronounced bend discontinuities. This blending can effectively smoothen out the output. Instead of using (u,v) as the coordinates in the circular disc for mapping, we introduce a new blended point ($u_{blend}$,$v_{blend}$) as output in lieu of (u,v). This point ($u_{blend}$,$v_{blend}$) is calculated by doing linear blending between (u,v) and (x,y). The main equation for this is

$$\begin{bmatrix} u_{blend} \\ v_{blend} \end{bmatrix} = \tau \begin{bmatrix} u \\ v \end{bmatrix} + (1-\tau) \begin{bmatrix} x \\ y \end{bmatrix}$$

where $\tau$ is the mapping blend factor defined as $\tau = (u^2 + v^2)^\rho$ with parameter ρ. This blend factor is a power of the distance between point (u,v) and the center of the circle. It tends to bias the blending between (u,v) and (x,y) towards the point (x,y) near the center of the circle. Similarly, it tends to bias the blending towards the point (u,v) near the perimeter of the square. The extra variable ρ is a user-provided parameter for artistic control of the roundness of the mapping. Setting ρ=0 turns off the blending and reverts this mapping to the simple stretching map. Setting ρ to 1 or 2 makes the mapping circular near the center and less-rounded near the perimeter. In effect, the parameter ρ allows the user to vary the roundness of the result.

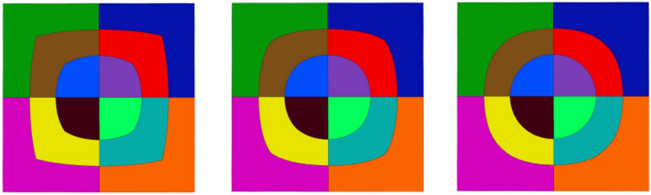

Figure A1.3: blending the simple stretching map with varying roundness at ρ=0.5 (left), ρ=1 (center), ρ=2 (right)

### A1.3 A Quasi-Equiareal Mapping Based on the Fernandez-Guasti Squircle

Fernandez-Guasti's squircle encloses an area A given by the equation:

$$A = \frac{4k^2 E(\sin^{-1} s, \frac{1}{s})}{s}$$

where $E(\varphi, k_E)$ is the Legendre elliptical integral of the 2nd kind. Using this formula on a unit squircle, we can design a quasi-equiareal radial mapping of a circular disc to a square. By quasi-equiareal, we mean that this mapping is close to being equiareal but not completely equiareal. It is actually easy to show that there are no radial mappings between the circle and the square that is equiareal.

For this mapping, we set:

$$s = \sqrt{x^2 + y^2 - x^2 y^2} \qquad t = \sqrt{s\, E(\sin^{-1} s, \frac{1}{s})}$$

and substitute *t* back to the radial constraint equations in Section 6.1 to get expressions for u and v in terms of x and y. This will give us a radially-constrained quasi-equiareal mapping based on the Fernandez-Guasti squircle. This mapping makes an adjustment on the FG-squircular mapping so that circular contours inside the circle have proportionally the same area as FG-squircles inside the square.

$$u = \sqrt{s\, E(\sin^{-1} s, \frac{1}{s})} \frac{x}{\sqrt{x^2 + y^2}}$$

$$v = \sqrt{s\, E(\sin^{-1} s, \frac{1}{s})} \frac{y}{\sqrt{x^2 + y^2}}$$

Note: For the sake of brevity in the equations, we have avoided expanding the common subexpression for *s* in terms of x and y.

This quasi-equiareal mapping gives results very similar to the FG-squircular mapping discussed in Section 6. However, it is much more computationally expensive because it requires the evaluation of Legendre elliptic function at every pixel. Hence, we recommend using the FG-squircular mapping over this quasi-equiareal mapping unless it is desirable to minimize equiareal error. See Figure A1.4 for a side-by-side comparison of the results between the FG-squircular mapping and this quasi-equiareal squircular mapping.

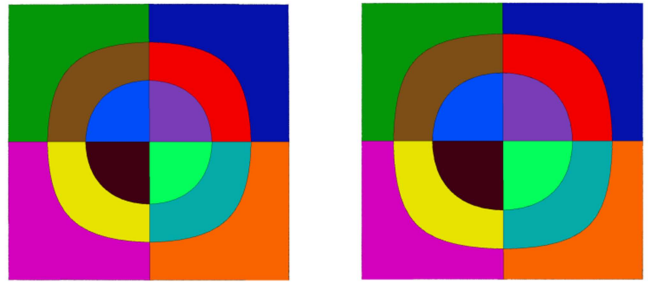

Figure A1.4: A comparison of results: FG-squircular mapping (left) and quasi-equiareal squircular mapping (right)

There is actually a way to further adjust this mapping to make it equiareal. However, this modification will alter the mapping so that it is no longer radial. This involves computing the incomplete arc length of the Fernandez-Guasti squircle and making adjustments in the mapping such that circular arc lengths inside the disc are proportional to squircular arc lengths inside the square. Unfortunately, we have not been able to find nice closed-form analytical expressions for this sort of mapping.

# Appendix 2: An Analogous Result for Blending Cylindrical Map Projections

Blending between conformal and equiareal map projections is not limited to the azimuthal family of map projections. In this appendix, we shall give a closed-form analytical expression for blending between conformal and equiareal cylindrical projections. In fact, we believe that for certain well-defined families of map projections, there exists a blending continuum between conformal and equiareal projections. We call this continuum as the *conformal-equiareal spectrum.*

In the main part of this paper, we showed a formula for a continuum of azimuthal projections between the conformal stereographic projection and the equiareal Lambert azimuthal projection. Similarly, in this appendix, we would like to present a formula for a continuum of cylindrical projections between the conformal Mercator projection and the equiareal Lambert cylindrical projection.

First, let us review what constitutes a cylindrical map projection. In its bare essence, the equations for a cylindrical map projection are:

$$x = k\lambda$$
$$y = f(\phi)$$

where *k* is a constant and *f* is a specific function.

## A2.1 The Mercator projection

The Mercator projection is probably the most famous cylindrical map projection of the world. It was discovered by Gerardus Mercator in 1569. However, it was not mathematically analyzed until the time of Edward Wright in 1599. The equations for this conformal projection are:

$$x = \lambda \qquad y = \ln[\tan\left(\frac{\pi}{4} + \frac{\phi}{2}\right)]$$

It is important to note that Mercator projection projects the sphere to an infinitely long strip with $x \epsilon [-\pi, \pi]$ and $y \epsilon (-\infty, +\infty)$.

## A2.2 The Lambert Cylindrical Equal-Area Projection

In his seminal 1772 treatise on map projections: "Notes and Comments on the Composition of Terrestrial and Celestial Maps", Johann Heinrich Lambert used techniques from calculus to develop an equiareal cylindrical map projection. The equations for this projection are:

$$x = \lambda \qquad y = \sin\phi$$

Unlike the Mercator, the Lambert cylindrical equal area projection has finite bounds with $x \epsilon [-\pi, \pi]$ and $y \epsilon [-1, 1]$.

## A2.3 A Blended Cylindrical Map Projection

Our proposed blended cylindrical map projection is:

$$y = sgn(\phi)\, \frac{1+\beta}{2\beta} \left[ \left(\frac{1 - \beta + \sin|\phi|}{1 - (1-\beta)\sin|\phi|}\right)^{\beta} - (1-\beta)^{\beta} \right]$$

Its inverse is:

$$\phi = sgn(y)\, \sin^{-1}\left( \frac{\left[(1-\beta)^{\beta} + |y|\frac{2\beta}{1+\beta}\right]^{\frac{1}{\beta}} - 1 + \beta}{\left[(1-\beta)^{\beta} + |y|\frac{2\beta}{1+\beta}\right]^{\frac{1}{\beta}}(1-\beta) + 1} \right)$$

It can be checked that at the limit when β approaches 0, the blended cylindrical map becomes the Mercator projection. Likewise, when β approaches 1, the blended cylindrical map becomes the Lambert cylindrical equal area projection.

Note that we're using a common mathematical function known as the signum function in our proposed equations. The signum function is defined as:

$$sgn(x) = \frac{|x|}{x} = \begin{cases} -1 & \text{if } x < 0, \\ 0 & \text{if } x = 0, \\ 1 & \text{if } x > 0. \end{cases}$$

## A2.2 Some Properties

Our proposed cylindrical blending function has some important properties that we would like to discuss here.

1) Equatorial Constraint for f(φ)

Both the Mercator and the Lambert cylindrical equal area projections have a value of y=0 when φ=0. Ideally, we want the blended cylindrical projection to maintain this property. It is easy to check that this is the case for the equations provided.

2) Symmetry: f(-φ) = - f(φ)

In other words, we want f(φ) to be an odd function. This property is important so that the value of y at the southern limit when φ= -90° is the just the negative of the value at the northern limit when φ=90°.

3) Finite

We want y to have a finite value except for the limiting case for Mercator when β=0 and φ=±90° where y is infinite. In other words, the blended cylindrical projection is finite for all φ. Only the Mercator projection is allowed to have an infinite value at the poles.

4) Alternative to the Equirectangular Projection

We can use a blend value of β=0.460711 to produce a blended cylindrical projection with 2:1 ratio of width versus height. This projection can effectively act as an alternative to the equirectangular projection in presenting to the sphere as a cylindrical projection with a compromise between being conformal and being equiareal.

## A2.4 Generalized Cylindrical Equal Areal Projection

Lambert's cylindrical equal area projection is not unique as an equiareal cylindrical projection. In fact, there is a whole class of cylindrical equal area projections that are stretched variations of Lambert's projection. The amount of stretching is facilitated by a parameter $\phi_0$. This parameter is called the standard latitude of cylindrical map projection. The equations for the generalized cylindrical equal area projection are:

$$x = \lambda \cos\phi_0 \qquad y = \frac{\sin\phi}{\cos\phi_0}$$

When $\phi_0 = 0°$, the generalized cylindrical equal area projection is just Lambert's projection. Other examples of specific values for $\phi_0$ include: Behrmann's projection at $\phi_0 = 30°$, Gall-Peters projection at $\phi_0 = 45°$, and Tobler's world in a square at $\phi_0 = 55.65°$.

Like the specific case for Lambert's cylindrical projection, the generalized cylindrical equal area projection is finite. It encompasses a rectangle in the Cartesian plane with the bounds $x \epsilon [-\pi \cos\phi_0, \pi \cos\phi_0]$ and $y \epsilon [-\frac{1}{\cos\phi_0}, \frac{1}{\cos\phi_0}]$.

We can easily extend our blended cylindrical map projection to have a parameter $\phi_0$ to blend between the conformal Mercator and the generalized cylindrical equal area projection. The equations for blending are summarized in Tables A2.1 and A2.2

| Cylindrical Projection | $y = f(\phi)$ | $\phi = f^{-1}(y)$ | key property | mapping span | blend value |
|---|---|---|---|---|---|
| Mercator | $\ln[\tan\left(\frac{\pi}{4} + \frac{\phi}{2}\right)]$ | $\frac{\pi}{2} - 2\tan^{-1}(e^{-y})$ | conformal | $-\infty < y < \infty$ | 0 |
| Lambert | $\sin\phi$ | $\sin^{-1} y$ | equiareal | $-1 \leq y \leq 1$ | 1 |
| blended | $sgn(\phi)\frac{1+\beta}{2\beta}\left[\left(\frac{1-\beta+\sin|\phi|}{1-(1-\beta)\sin|\phi|}\right)^\beta - (1-\beta)^\beta\right]$ | $sgn(y)\sin^{-1}\left(\frac{\left[(1-\beta)^\beta + |y|\frac{2\beta}{1+\beta}\right]^{\frac{1}{\beta}} - 1 + \beta}{\left[(1-\beta)^\beta + |y|\frac{2\beta}{1+\beta}\right]^{\frac{1}{\beta}}(1-\beta) + 1}\right)$ | adjustable | grows as $\beta \to 0$ | $\beta$ |

Table A2.1: Blending the Mercator and Lambert cylindrical equal area projections

| Cylindrical Projection | $y = f(\phi)$ | $\phi = f^{-1}(y)$ | Longitude equation |
|---|---|---|---|
| Mercator | $\ln[\tan\left(\frac{\pi}{4} + \frac{\phi}{2}\right)]$ | $\frac{\pi}{2} - 2\tan^{-1}(e^{-y})$ | $x = \lambda$ |
| Generalized cylindrical equal area | $\frac{\sin\phi}{\cos\phi_0}$ | $\sin^{-1}(y\cos\phi_0)$ | $x = \lambda\cos\phi_0$ |
| blended | $sgn(\phi)\frac{1+\beta}{2\beta}\left[\left(\frac{1-\beta+\sin|\phi|}{1-(1-\beta)\sin|\phi|}\right)^\beta - (1-\beta)^\beta\right]\left(\frac{1}{\cos\phi_0}\right)^\beta$ | $sgn(y)\sin^{-1}\left(\frac{\left[(1-\beta)^\beta + |y|\frac{2\beta}{1+\beta}\right]^{\frac{1}{\beta}} - 1 + \beta}{\left[(1-\beta)^\beta + |y|\frac{2\beta}{1+\beta}\right]^{\frac{1}{\beta}}(1-\beta) + 1}\right)(\cos\phi_0)^\beta$ | $x = \lambda(\cos\phi_0)^\beta$ |

Table A2.2: Blending the Mercator and the generalized cylindrical equal area projection.

Notes for the derivation of this blended mapping:

1) The Mercator projection has alternative equations in the form of

$$y = \frac{1}{2}\ln\left(\frac{1+\sin\phi}{1-\sin\phi}\right)$$

and

$$y = \ln\left(\frac{1+\sin\phi}{\cos\phi}\right)$$

2) The following limit is true for the natural logarithm

$$\lim_{\beta \to 0} \frac{1}{\beta}\left(x^\beta - (1-\beta)^\beta\right) = \ln x$$

3) This proposed blended cylindrical projection is by no means unique. There are other ways to blend the Mercator and the Lambert cylindrical equal area projection. For example, this projection will also work:

$$y = \frac{sgn(\phi)}{\beta}\left[\left(\frac{1-\beta+\sin|\phi|}{\sqrt{1-(1-\beta^2)\sin^2\phi}}\right)^\beta - (1-\beta)^\beta\right]$$

However, the inverse equation for this projection is particularly gnarly.

14